\documentclass{article}

\usepackage{etoolbox}
\usepackage[preprint]{neurips_2025}

\newif\ifpreprintorfinal

\makeatletter
\@ifpackagewith{neurips_2025}{preprint}{\preprintorfinaltrue}{}
\@ifpackagewith{neurips_2025}{final}{\preprintorfinaltrue}{}
\makeatother

\newcommand{\anon}[2]{%
    \ifpreprintorfinal
        #1%
    \else
        #2%
    \fi
}

\usepackage{subcaption}
\usepackage{hyperref}
\usepackage{multirow}
\usepackage{booktabs}
\usepackage{placeins}
\usepackage{amsmath}
\usepackage{array}
\usepackage{tikz}

\usepackage{pgfplots}
\usepgfplotslibrary{dateplot}
\pgfplotsset{
    compat=newest,
    every axis/.append style={
        width=10cm,
        height=6cm,
        font=\small,
        tick pos=left,
        major grid style={
            dashed,
            gray!30
        },
        grid=both
    }
}

\newcommand{\wordheat}[2]{%
    \begingroup
        \pgfmathsetmacro{\val}{#1}%
        \ifdim \val pt < 0 pt
          \pgfmathsetmacro{\t}{min(1, max(0, -\val))}%
          \pgfmathsetmacro{\r}{(1 - \t) + \t * (230/255)}%
          \pgfmathsetmacro{\g}{(1 - \t) + \t * (25/255)}%
          \pgfmathsetmacro{\b}{(1 - \t) + \t * (75/255)}%
        \else
          \pgfmathsetmacro{\t}{min(1, max(0, \val))}%
          \pgfmathsetmacro{\r}{(1 - \t) + \t * (60/255)}%
          \pgfmathsetmacro{\g}{(1 - \t) + \t * (180/255)}%
          \pgfmathsetmacro{\b}{(1 - \t) + \t * (75/255)}%
        \fi
        \definecolor{tempcolor}{rgb}{\r, \g, \b}%
        \tikz[baseline=(W.base)]{%
          \node[
            inner sep=0pt,
            outer sep=0pt,
            anchor=base,
            fill=tempcolor,
            text=black
          ] (W) {#2};%
        }%
    \endgroup\allowbreak
}

\usepackage{xcolor}
\definecolor{color1}{HTML}{108810}      
\definecolor{color2}{HTML}{0000FF}      
\definecolor{color3}{HTML}{FF00FF}      
\definecolor{color4}{HTML}{FF0000}      
\definecolor{color5}{HTML}{FF8C00} 

\title{ForeCite: Adapting Pre-Trained Language Models to Predict Future Citation Rates of Academic Papers}

\author{
    Gavin Hull\\
    Department of Computer Science, Department of Mathematics and Statistics\\
    Memorial University of Newfoundland and Labrador\\
    St. John's, NL A1C 5S7\\
    \texttt{ghull@mun.ca}\\
    \And
    Alex Bihlo\\
    Department of Mathematics and Statistics\\
    Memorial University of Newfoundland and Labrador\\
    St. John's, NL A1C 5S7\\
    \texttt{abihlo@mun.ca}
}

\begin{document}

\maketitle

\begin{abstract}
    Predicting the future citation rates of academic papers is an important step toward the automation of research evaluation and the acceleration of scientific progress. We present \textbf{ForeCite}, a simple but powerful framework to append pre-trained causal language models with a linear head for average monthly citation rate prediction. Adapting transformers for regression tasks, ForeCite achieves a test correlation of $\rho = 0.826$ on a curated dataset of 900K+ biomedical papers published between 2000 and 2024, a 27-point improvement over the previous state-of-the-art. Comprehensive scaling-law analysis reveals consistent gains across model sizes and data volumes, while temporal holdout experiments confirm practical robustness. Gradient-based saliency heatmaps suggest a potentially undue reliance on titles and abstract texts. These results establish a new state-of-the-art in forecasting  the long-term influence of academic research and lay the groundwork for the automated, high-fidelity evaluation of scientific contributions.\footnote{Relevant code will be made available upon publication.}
\end{abstract}

\section{Introduction}

This paper explores the potential of pre-trained language models to predict future citation rates directly from the textual content of scientific manuscripts. Though citation rates serve only as a weak proxy for true scholarly impact, they offer a tractable signal through which to examine how contemporary models might begin to interpret and anticipate patterns of scientific attention at scale. By framing citation prediction as a text-to-signal learning problem, this work establishes a foundation for more context-aware, adaptive systems that could assist in analyzing\textemdash and eventually guiding\textemdash the evolving landscape of scientific research.

Traditional citation-prediction methods depend heavily on external metadata\textemdash author profiles, citation graphs, or peer-review excerpts\textemdash to anticipate future citations \citep{pobiedina2016citation, davletov2014high, li-etal-2019-neural}. While these features can yield reasonable forecasts in narrow domains, they inherently overlook the semantic information and methodological quality embedded in the manuscript itself. In contrast, modern language models have been shown to capture the deep contextual and conceptual relationships in text \citep{brown2020languagemodelsfewshotlearners}, although their application to citation forecasting has largely been limited to small text fragments or as embeddings for separate downstream learners \citep{hirako2024cimatecitationcountprediction, hirako-etal-2023-realistic, van_Dongen_2020, wenniger2023multischubert}.

To bridge this gap, we introduce ForeCite, a plug-and-play framework for adapting pre-trained causal transformers for regression tasks, enabling end-to-end prediction of average monthly citation rates. We evaluate ForeCite on a curated corpus of 900K+ medical articles publish between 2000 and 2024, demonstrating superior predictive performance over state-of-the-art baselines. Our extensive suite of experiments (including scaling-law analysis across model sizes and data volumes, temporal holdout studies to probe concept drift, gradient-based saliency visualizations for interpretability, and targeted ablations) evidence the potential of large language models (LLMs) in scientific impact forecasting.

We make four main contributions:
\begin{enumerate}
    \item \textbf{Framework for regression.} A simple general method for adapting pre-trained causal language models for regression tasks while preserving semantic knowledge.
    \item \textbf{New state-of-the-art.} A Spearman rank correlation of $\rho = 0.826$ between predicted and true citation rates, marking a 27-point improvement over the previous state-of-the-art.
    \item \textbf{Comprehensive testing.} In-depth analysis of scaling behavior, temporal robustness, interpretability, and targeted ablations.
    \item \textbf{Open science commitment.} Public release of code and model checkpoints of various sizes to spur further advancements in this field.
\end{enumerate}

\section{Related work}

\paragraph{Citation prediction.}  Early work framed citation forecasting as a classification or simple regression problem, using handcrafted features derived from author and venue metadata \citep{ibanez2009predicting, chakraborty2014towards, pobiedina2016citation}. Alternative graph-based methods exploited network structure, both temporal and topological, to extend forecasting horizons \citep{davletov2014high}. Some efforts also employed manual mathematical modeling to capture citation dynamics more explicitly \citep{wang2013quantifying}.

More recent neural models incorporate textual and early-citation signals via multilayer perceptrons or recurrent neural networks, yielding improved but still limited performance \citep{ruan2020predicting, jamal2024forecasting}. The latest transformer-based predictors leverage pre-trained language embeddings over abstracts or document chunks, but typically only use them as fixed features for downstream regressors rather than in an end-to-end predictive framework \citep{hirako2024cimatecitationcountprediction, van_Dongen_2020, hirako-etal-2023-realistic, wenniger2023multischubert}. A more in-depth analysis of techniques used in the field can be found in \citet{aiza2024features}.

\paragraph{Large language models.} The transformer architecture, built solely on self-attention and forgoing recurrence, has become a foundational tool in  natural language processing (NLP), demonstrating superior performance and parallelizability across tasks \citep{vaswani2023attentionneed}. Scaling transformers to billions of parameters has yielded emergent capabilities, including few-shot learning, that have redefined model generality and adaptability \citep{brown2020languagemodelsfewshotlearners}. Despite these advances, fine-tuning causal transformers for regression tasks such as citation rate prediction remains largely underexplored in the literature.

\paragraph{Scaling-laws.} Empirical studies reveal that language models' performance follows predictable power-law relationships with respect to both model size and training data volume \citep{kaplan2020scalinglawsneurallanguage}. While these scaling-laws extend to many downstream tasks, the precise improvement achievable through fine-tuning is task dependent. Our work is the first to systematically characterize these trends for citation rate prediction.

\paragraph{Interpretability.} Understanding which textual elements drive model predictions is critical for transparency and interpretability. Interpretability techniques for transformer-based architectures fall into several categories. Attention-based visualization remains a popular means of attributing model decisions, highlighting tokens which receive high self-attention scores \citep{hao2021selfattentionattributioninterpretinginformation}. Gradient-based saliency methods, originally developed for vision models and later adapted for NLP, assign importance scores to input tokens via back-propagated gradients \citep{li2016visualizingunderstandingneuralmodels}. We apply these techniques here to highlight the impact of individual tokens on predicted citation rates.

\section{Methods}
\subsection{Data collection \& preprocessing}

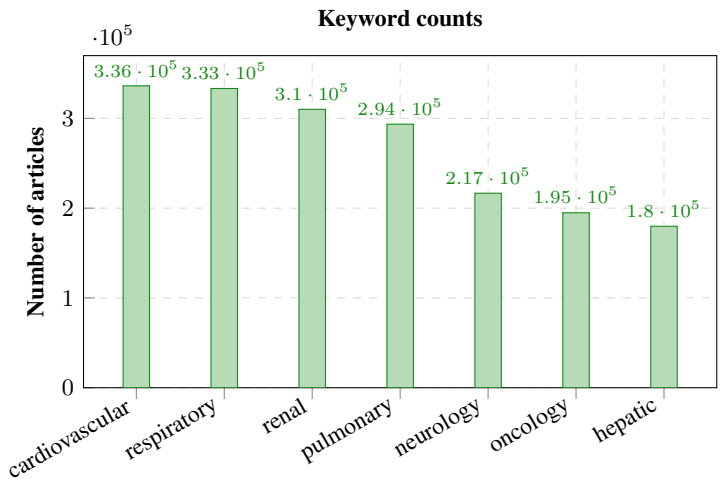
\begin{figure}[!ht]
    \centering

    \begin{tikzpicture}
        \begin{axis}[
            ybar,
            symbolic x coords={cardiovascular,respiratory,renal,pulmonary,neurology,oncology,hepatic},
            xtick=data,
            x tick label style={
                rotate=30,
                anchor=east
            },
            ylabel={\textbf{Number of articles}},
            ymin=0,
            title={\textbf{Keyword counts}},
            nodes near coords,
            every node near coord/.append style={
                font=\scriptsize,
                text=color1
            }
        ]

            \addplot+[
                ybar,
                draw=color1,
                fill=color1!30,
            ] coordinates {
                (cardiovascular,336212)
                (respiratory,333282)
                (renal,310085)
                (pulmonary,293531)
                (neurology,216621)
                (oncology,194859)
                (hepatic,179839)
            };
            
        \end{axis}
    \end{tikzpicture}

    \caption{Distribution of articles in the final dataset containing each domain-specific keyword.}
    \label{fig:keyword-counts}
\end{figure}

We constructed our dataset by querying the Elsevier API for seven biomedical keywords across every month from January 2000 to December 2024 with a maximum of 6K resulting articles per query. After de-duplication, our dataset underwent rigorous filtering of corrupted entries, non-English and restricted-access items, and outliers via several criteria (e.g. article length, completeness, etc.). This process yielded a high-quality corpus comprising 900K+ articles (see Figure~\ref{fig:keyword-counts}), each annotated with its publication date and total number of citations as of December 2024.

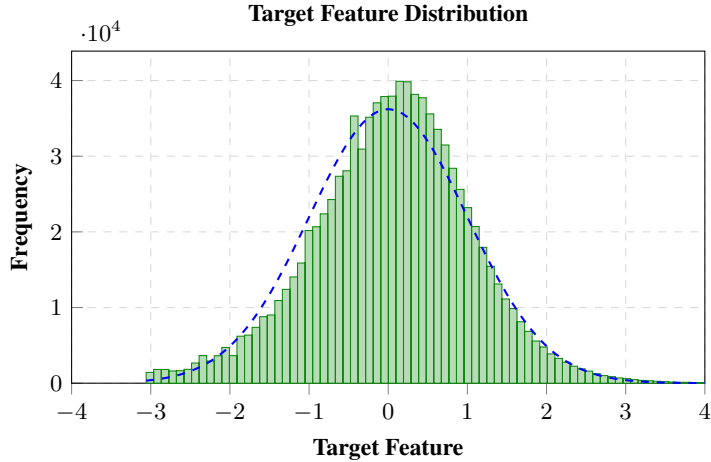
\begin{figure}[!ht]
    \centering

    \begin{tikzpicture}
        \begin{axis}[
            ybar,
            xmin=-4,
            xmax=4,
            xlabel={\textbf{Target Feature}},
            ylabel={\textbf{Frequency}},
            ymin=0      ,
            title={\textbf{Target Feature Distribution}}
        ]

            \addplot+[
                ybar interval,
                mark=no,
                fill=color1!30,
                draw=color1
            ] coordinates {
                (-3.05853, 1427)
                (-2.96290, 1812)
                (-2.86727, 1815)
                (-2.77164, 1612)
                (-2.67602, 1655)
                (-2.58039, 1835)
                (-2.48476, 2664)
                (-2.38913, 3649)
                (-2.29350, 2924)
                (-2.19787, 3633)
                (-2.10224, 4718)
                (-2.00661, 3648)
                (-1.91098, 6210)
                (-1.81536, 6356)
                (-1.71973, 7383)
                (-1.62410, 8790)
                (-1.52847, 9019)
                (-1.43284, 10929)
                (-1.33721, 12404)
                (-1.24158, 14044)
                (-1.14595, 15882)
                (-1.05032, 20176)
                (-0.95469, 20677)
                (-0.85907, 22382)
                (-0.76344, 24272)
                (-0.66781, 27346)
                (-0.57218, 28073)
                (-0.47655, 35319)
                (-0.38092, 30937)
                (-0.28529, 35148)
                (-0.18966, 37055)
                (-0.09403, 37883)
                (0.00159, 37930)
                (0.09722, 39899)
                (0.19285, 39836)
                (0.28848, 38173)
                (0.38411, 37727)
                (0.47974, 35591)
                (0.57537, 33551)
                (0.67100, 31482)
                (0.76663, 28402)
                (0.86225, 25618)
                (0.95788, 23202)
                (1.05351, 20703)
                (1.14914, 17953)
                (1.24477, 15450)
                (1.34040, 13114)
                (1.43603, 11129)
                (1.53166, 9868)
                (1.62729, 8124)
                (1.72292, 6850)
                (1.81854, 5575)
                (1.91417, 4783)
                (2.00980, 3871)
                (2.10543, 3277)
                (2.20106, 2776)
                (2.29669, 2245)
                (2.39232, 1898)
                (2.48795, 1592)
                (2.58358, 1246)
                (2.67920, 1005)
                (2.77483, 862)
                (2.87046, 695)
                (2.96609, 601)
                (3.06172, 472)
                (3.15735, 372)
                (3.25298, 323)
                (3.34861, 254)
                (3.44424, 203)
                (3.53987, 157)
                (3.63549, 114)
                (3.73112, 115)
                (3.82675, 89)
                (3.92238, 69)
                (4.01801, 40)
            };
    
            \addplot+[
                smooth,
                draw=color2,
                dashed,
                thick,
                fill=none,
                forget plot,
                no marks
            ] coordinates {
                (-3.05853, 336.88157)
                (-2.96290, 449.27859)
                (-2.86727, 593.72127)
                (-2.77164, 777.45975)
                (-2.67602, 1008.79204)
                (-2.58039, 1297.04130)
                (-2.48476, 1652.47303)
                (-2.38913, 2086.13967)
                (-2.29350, 2609.64123)
                (-2.19787, 3234.79416)
                (-2.10224, 3973.20451)
                (-2.00661, 4835.74745)
                (-1.91098, 5831.96254)
                (-1.81536, 6969.38190)
                (-1.71973, 8252.81715)
                (-1.62410, 9683.63934)
                (-1.52847, 11259.09310)
                (-1.43284, 12971.69240)
                (-1.33721, 14808.74620)
                (-1.24158, 16752.06560)
                (-1.14595, 18777.89310)
                (-1.05032, 20857.09320)
                (-0.95469, 22955.62510)
                (-0.85907, 25035.30520)
                (-0.76344, 27054.84670)
                (-0.66781, 28971.14690)
                (-0.57218, 30740.76860)
                (-0.47655, 32321.54990)
                (-0.38092, 33674.25980)
                (-0.28529, 34764.20960)
                (-0.18966, 35562.72930)
                (-0.09403, 36048.41980)
                (0.00159, 36208.10550)
                (0.09722, 36037.42870)
                (0.19285, 35541.04650)
                (0.28848, 34732.42050)
                (0.38411, 33633.20950)
                (0.47974, 32272.30580)
                (0.57537, 30684.57440)
                (0.67100, 28909.37040)
                (0.76663, 26988.92500)
                (0.86225, 24966.68960)
                (0.95788, 22885.72950)
                (1.05351, 20787.24730)
                (1.14914, 18709.30370)
                (1.24477, 16685.78690)
                (1.34040, 14745.65890)
                (1.43603, 12912.49290)
                (1.53166, 11204.29230)
                (1.62729, 9633.56845)
                (1.72292, 8207.64131)
                (1.81854, 6929.11824)
                (1.91417, 5796.50211)
                (2.00980, 4804.87891)
                (2.10543, 3946.63824)
                (2.20106, 3212.18548)
                (2.29669, 2590.61176)
                (2.39232, 2070.29614)
                (2.48795, 1639.42304)
                (2.58358, 1286.40589)
                (2.67920, 1000.21515)
                (2.77483, 770.61465)
                (2.87046, 588.31445)
                (2.96609, 445.05142)
                (3.06172, 333.61018)
                (3.15735, 247.79744)
                (3.25298, 182.38233)
                (3.34861, 133.01393)
                (3.44424, 96.12579)
                (3.53987, 68.83529)
                (3.63549, 48.84395)
                (3.73112, 34.34305)
                (3.82675, 23.92739)
                (3.92238, 16.51887)
                (4.01801, 11.30039)
            };
            
        \end{axis}
    \end{tikzpicture}

    \caption{Standardized average monthly citation rates post log-transform, overlaid with a scaled standard normal curve for reference.}
    \label{fig:target-histogram}
\end{figure}

Aligning with standard practice in citation prediction \citep{hirako2024cimatecitationcountprediction, van_Dongen_2020, hirako-etal-2023-realistic, wenniger2023multischubert}, we applied a log-transform to the average monthly citation rates. This produced a near-Gaussian distribution (Figure~\ref{fig:target-histogram}), motivating the standardization of our data prior to training. The mean and standard deviation were calculated using the training split to prevent leakage. Given the large quantities of parameters aggregated during LLM inference, the central limit theorem suggests that this transformation should encourage more stable optimization.

Each article, originally in XML format, was converted to a consistent Markdown representation via a custom XSLT script. This standardized format preserved structural elements such as section headings, section content, and figure/table captions, as well as available metadata such as journal name and copyright, enabling consistent tokenization across all models.

\FloatBarrier

\subsection{Model architecture}

All experiments followed an identical architectural pipeline: we appended a simple linear head to the final hidden state of a pre-trained Hugging Face \texttt{AutoModelForCausalLM} model serving as the base language model. All other architectural decisions (e.g. tokenization scheme, context window length, etc.) were inherited from the base model configurations. To maximize hardware utilization, we apply 4-bit NF4 quantization and \texttt{bfloat16} arithmetic; this configuration enabled highly efficient training, allowing even our largest models to fit within our memory budget without sacrificing predictive fidelity.

\subsection{Training procedure}

Each model was trained in two phases:
\begin{enumerate}
    \item \textbf{Initial training phase:} The base LLM was frozen while only the appended linear head was trained. 
    \item \textbf{Fine-tuning phase:} The base LLM was unfrozen and jointly fine-tuned with the linear head via QLoRA \citep{dettmers2023qloraefficientfinetuningquantized}.
\end{enumerate}

All training runs used a 90/10 train/test split. A complete list of training and fine-tuning hyperparameters is provided in Table~\ref{tab:hyperparams} in Appendix~\ref{app:hyperparams}. Any parameter not listed was left at its default value.

All experiments were conducted on Compute Canada's \textit{Narval} cluster, using $6 \times$ AMD EPYC 7413 CPUs and $2 \times$ NVIDIA A100 SXM4 GPUs. Although each run's wall-clock time varied by model size and data volume, we exclude these details here for brevity. All experiments had a maximum run-time of 7 days.

\subsection{Experimental setup}

In the section that follows, we present the results of our two primary experiments:
\paragraph{Scaling-law analysis.} To characterize how the performance of our framework scales with model size and data volume, we trained five transformer variants (0.5B \textendash 14B parameters) across five fractions of our corpus (1\% \textendash 16\%). Data volumes never exceed 16\% of our corpus due to computational constraints. A full list of the models used in this work, alongside their publishers and parameter counts, can be found in Appendix~\ref{app:models}, Table~\ref{tab:models}. All evaluations were\textemdash unless otherwise specified\textemdash performed on a randomly generated hold-out test set. A full list of our evaluation metrics can be found in Appendix~\ref{app:metrics}. 

\paragraph{Gradient saliency analysis.} To reveal which textual features drive ForeCite's predictions, we computed gradient-based attributions over a number of samples \citep{li2016visualizingunderstandingneuralmodels}. For each sample, we accumulated token-level gradients of the predicted score with respect to the input embeddings, quantifying the impact of individual tokens on predicted citations.

\paragraph{Temporal holdout.} To verify the robustness of our framework in real-world applications, we trained Bloom-560m on a random subset of articles published before 2023 (equivalent to 16\% of our total corpus). Evaluating on monthly cohorts thereafter, we reported a rolling Pearson correlation. The Bloom-560m model was selected due to its release date of 2022, ensuring that no test data could have been seen during its pre-training.

\section{Results}

\subsection{Scaling-law analysis}

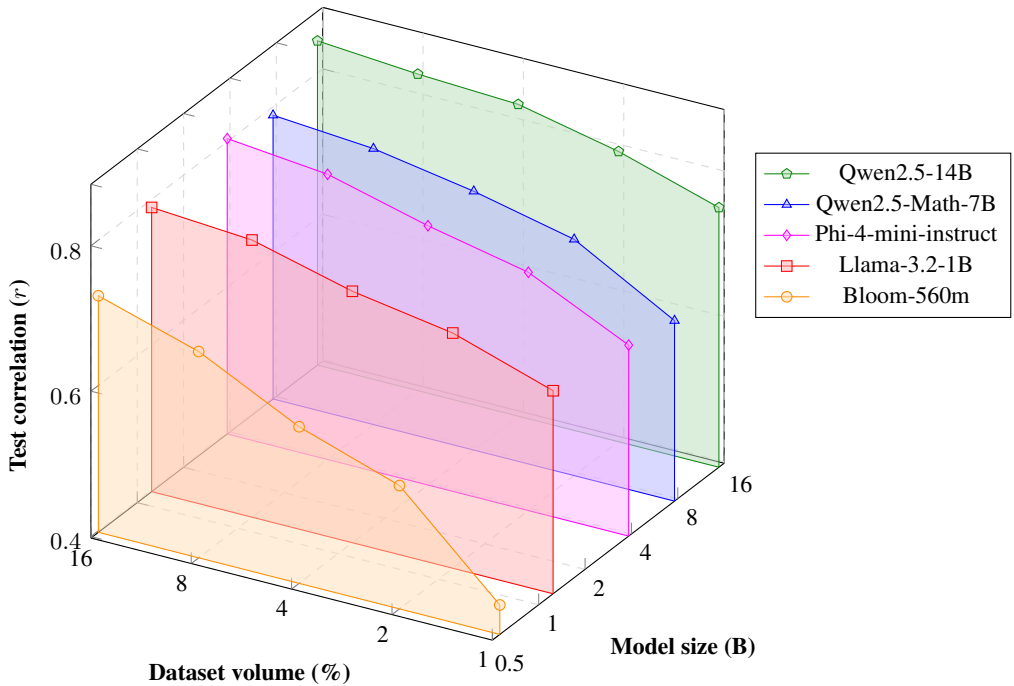
\begin{figure}[!ht]
    \centering
    
    \begin{tikzpicture}
        \pgfplotstableread{
            0.16 0.559 0.7234
            0.08 0.559 0.6818
            0.04 0.559 0.6134
            0.02 0.559 0.5677
            0.01 0.559 0.4385
        }\modelA

        \pgfplotstableread{
            0.16 1.24 0.7890
            0.08 1.24 0.7793
            0.04 1.24 0.7439
            0.02 1.24 0.7216
            0.01 1.24 0.6773
        }\modelB

        \pgfplotstableread{
            0.16 3.84 0.8044
            0.08 3.84 0.7904
            0.04 3.84 0.7547
            0.02 3.84 0.7258
            0.01 3.84 0.6607
        }\modelC

        \pgfplotstableread{
            0.16 7.62 0.7884
            0.08 7.62 0.7776
            0.04 7.62 0.7540
            0.02 7.62 0.7230
            0.01 7.62 0.6463
        }\modelD

        \pgfplotstableread{
            0.16 14.8 0.8441
            0.08 14.8 0.8335
            0.04 14.8 0.8270
            0.02 14.8 0.7973
            0.01 14.8 0.7552
        }\modelE

        \begin{axis}[
            view={30}{30},
            x dir=reverse,
            height=10cm,
            xlabel={\textbf{Dataset volume (\%)}},
            xmode=log,
            log basis x=2,
            xtick={0.16,0.08,0.04,0.02,0.01},
            xticklabels={16,8,4,2,1},
            ylabel={\textbf{Model size (B)}},
            ymode=log,
            ymin=0.5,
            ymax=16,
            ytick={0.5,1,2,4,8,16},
            yticklabels={0.5,1,2,4,8,16},
            log basis y=2,
            zlabel={\textbf{Test correlation ($r$)}},
            legend style={
                at={(rel axis cs:0.5,2)},
                anchor=west
            },
            title={\textbf{3D scaling-laws: model size vs. data volume vs. test correlation ($r$)}}
        ]

            \addplot3[
                color=color1,
                draw=color1,
                fill=color1!30,
                fill opacity=0.5,
                mark=pentagon*
            ] table [x index=0, y index=1, z index=2] {\modelE} \closedcycle;
            \addlegendentry{Qwen2.5-14B}

            \addplot3[
                color=color2,
                draw=color2,
                fill=color2!30,
                fill opacity=0.5,
                mark=triangle*
            ] table [x index=0, y index=1, z index=2] {\modelD} \closedcycle;
            \addlegendentry{Qwen2.5-Math-7B}

            \addplot3[
                color=color3,
                draw=color3,
                fill=color3!30,
                fill opacity=0.5,
                mark=diamond*
            ] table [x index=0, y index=1, z index=2] {\modelC} \closedcycle;
            \addlegendentry{Phi-4-mini-instruct}

            \addplot3[
                color=color4,
                draw=color4,
                fill=color4!30,
                fill opacity=0.5,
                mark=square*
            ] table [x index=0, y index=1, z index=2] {\modelB} \closedcycle;
            \addlegendentry{Llama-3.2-1B}

            \addplot3[
                color=color5,
                draw=color5,
                fill=color5!30,
                fill opacity=0.5,
                mark=*
            ] table [x index=0, y index=1, z index=2] {\modelA} \closedcycle;
            \addlegendentry{Bloom-560m}
        \end{axis}
    \end{tikzpicture}

    \caption{A 3D visualization of the relationship between model size (in billions of parameters, $\log_2$ scaled), data volume (in percent of total corpus, $\log_2$ scaled), and performance (in Pearson correlation, evaluated on a holdout test set.)}
    \label{fig:scaling-3d}
\end{figure}

Figure~\ref{fig:scaling-3d} presents a 3D surface of Pearson correlation ($r$) over model sizes and data volumes. Our largest configuration, Qwen2.5-14B on 16\% data, attained $r = 0.844$, whereas our smallest, Bloom-560m on 1\% data, reported nearly half: $r = 0.430$. Figures~\ref{fig:model-size-vs-corr} and \ref{fig:data-vol-vs-corr} isolate these trends:
\begin{enumerate}
    \item As data volumes increase, all models improved monotonically, with steeper gains being recorded for the smaller models.
    \item As model size increases, performance generally rose, though mid-sized models (1B \textendash\ 8B) showed noticeable deviations.
\end{enumerate}

A complete grid of all 25 values for each of our evaluation metrics is provided in Appendix~\ref{app:scaling}, Table~\ref{tab:scaling}.

\begin{figure}[!ht]
    \centering
    \begin{subfigure}[b]{0.48\textwidth}
        \centering
        \begin{tikzpicture}
            \begin{axis}[
                width=\linewidth,
                height=7cm,
                xlabel={\textbf{Model size (B)}},
                xmode=log,
                log basis x=2,
                xtick={0.5,1,2,4,8,16},
                xticklabels={0.5,1,2,4,8,16},
                ylabel={\textbf{Test correlation ($r$)}},
                title={\textbf{Model size scaling-laws}},
                legend pos=south east,
                legend style={
                    nodes={
                        scale=0.8,
                        transform shape
                    }
                },
                legend cell align=left
            ]

                \addplot[color=black, mark=none, thick, dashed] coordinates {
                    (0.559, 0.810632568957)
                    (1.24,  0.836098473936)
                    (3.84,  0.866889828467)
                    (7.62,  0.882850665981)
                    (14.8,  0.896585819994)
                };
                \addlegendentry{100\% data$^*$}

                \addplot[color=color1!30, mark=pentagon*, mark options={color=color1}, thick] coordinates {
                    (0.559, 0.7234)
                    (1.24,  0.7890)
                    (3.84,  0.8044)
                    (7.62,  0.7884)
                    (14.8,  0.8441)
                };
                \addlegendentry{16\% data}

                \addplot[color=color2!30, mark=triangle*, mark options={color=color2}, thick] coordinates {
                    (0.559, 0.6818)
                    (1.24,  0.7793)
                    (3.84,  0.7904)
                    (7.62,  0.7776)
                    (14.8,  0.8335)
                };
                \addlegendentry{8\% data}

                \addplot[color=color3!30, mark=diamond*, mark options={color=color3}, thick] coordinates {
                    (0.559, 0.6134)
                    (1.24,  0.7439)
                    (3.84,  0.7547)
                    (7.62,  0.7540)
                    (14.8,  0.8270)
                };
                \addlegendentry{4\% data}

                \addplot[color=color4!30, mark=square*, mark options={color=color4}, thick] coordinates {
                    (0.559, 0.5677)
                    (1.24,  0.7216)
                    (3.84,  0.7258)
                    (7.62,  0.7230)
                    (14.8,  0.7973)
                };
                \addlegendentry{2\% data}

                \addplot[color=color5!30, mark=*, mark options={color=color5}, thick] coordinates {
                    (0.559, 0.4385)
                    (1.24,  0.6733)
                    (3.84,  0.6607)
                    (7.62,  0.6463)
                    (14.8,  0.7552)
                };
                \addlegendentry{1\% data}
            \end{axis}
        \end{tikzpicture}

        \caption{A 2D visualization of the relationship between model size (in billions of parameters, $\log_2$ scaled) and performance (in Pearson correlation, evaluated on a holdout test set) at various data volumes (in percent of total corpus.)}
        \label{fig:model-size-vs-corr}
    \end{subfigure}
    \hfill
    \begin{subfigure}[b]{0.48\textwidth}
        \centering
        \begin{tikzpicture}
            \begin{axis}[
                width=\linewidth,
                height=7cm,
                xlabel={\textbf{Data volume (\%)}},
                xmode=log,
                log basis x=2,
                xtick={0.01,0.02,0.04,0.08,0.16},
                xticklabels={1,2,4,8,16},
                ylabel={\textbf{Test correlation ($r$)}},
                title={\textbf{Data volume scaling-laws}},
                legend pos=south east,
                legend style={
                    nodes={
                        scale=0.8,
                        transform shape
                    }
                },
                legend cell align=left
            ]

                \addplot[color=black, mark=none, thick, dashed] coordinates {
                    (0.01, 0.845457829607)
                    (0.02, 0.865962874163)
                    (0.04, 0.883918390726)
                    (0.08, 0.899598021815)
                    (0.16, 0.913257255975)
                };
                \addlegendentry{Llama 4 Behemoth$^*$}
                
                \addplot[color=color1!30, mark=pentagon*, mark options={color=color1}, thick] coordinates {
                    (0.01, 0.7552)
                    (0.02, 0.7973)
                    (0.04, 0.8270)
                    (0.08, 0.8335)
                    (0.16, 0.8441)
                };
                \addlegendentry{Qwen2.5-14B}

                \addplot[color=color2!30, mark=triangle*, mark options={color=color2}, thick] coordinates {
                    (0.01, 0.6463)
                    (0.02, 0.7230)
                    (0.04, 0.7540)
                    (0.08, 0.7776)
                    (0.16, 0.7884)
                };
                \addlegendentry{Qwen2.5-Math-7B}

                \addplot[color=color3!30, mark=diamond*, mark options={color=color3}, thick] coordinates {
                    (0.01, 0.6607)
                    (0.02, 0.7258)
                    (0.04, 0.7547)
                    (0.08, 0.7904)
                    (0.16, 0.8044)
                };
                \addlegendentry{Phi-4-mini-instruct}

                \addplot[color=color4!30, mark=square*, mark options={color=color4}, thick] coordinates {
                    (0.01, 0.6773)
                    (0.02, 0.7216)
                    (0.04, 0.7439)
                    (0.08, 0.7793)
                    (0.16, 0.7890)
                };
                \addlegendentry{Llama-3.2-1B}

                \addplot[color=color5!30, mark=*, mark options={color=color5}, thick] coordinates {
                    (0.01, 0.4385)
                    (0.02, 0.5677)
                    (0.04, 0.6134)
                    (0.08, 0.6818)
                    (0.16, 0.7234)
                };
                \addlegendentry{Bloom-560m}
            \end{axis}
        \end{tikzpicture}

        \caption{A 2D visualization of the relationship between data volume (in percent of total corpus, $\log_2$ scaled) and performance (in Pearson correlation, evaluated on a holdout test set) at various model sizes (in billions of parameters.))}
        \label{fig:data-vol-vs-corr}
    \end{subfigure}

    \par\medskip
    \footnotesize{$^*$Indicates a theoretical estimate computed using our least squares model.}
    
\end{figure}

To anticipate the trajectory of state-of-the-art architectures trained on our full corpus, we fit a bounded least squares model of the form
\begin{equation}
    f(p, d) = \tanh(\beta_0 + \beta_1\log_2(p) + \beta_2\log_2 (d)),
\end{equation}
where $p$ is model size in billions of parameters, and $d$ is the percent of total data used. Extrapolating to the latest state-of-the-art Llama model, Llama 4 Behemoth, with 288 B active parameters and trained on our full corpus, we estimated a Pearson correlation of $r(288, 100) = 0.9413$ and  Spearman rank correlation of $\rho(288, 100) = 0.9325$. A full collection of the derived $\beta$-parameters can be found in Appendix~\ref{app:curve-fit}, Table~\ref{tab:curve-fit}. Both models exhibited an MAE of approximately 0.03.

\FloatBarrier

\subsection{Gradient saliency analysis}

On average, title and abstract tokens exhibited the highest attribution, with general body text receiving minimal weight. Across the sampled papers, models consistently favored focused and information-dense content that reflected the main topic of each paper. For example, "empty," meta-discursive statements such as "In this article we..." were routinely penalized, whereas discourse markers indicating new information such as "additionally" and "moreover," were often rewarded (albeit not as strongly as "empty" statements were penalized). Model attribution was inconsistent across most metadata\textemdash including publication dates and journal names\textemdash only consistently using author names for the most prominent authors. Figure~\ref{fig:gradient-saliency} in Appendix~\ref{app:gradient-saliency} displays a representative heatmap over the abstract of a sample paper \citep{arsenault2022does}.

\subsection{Temporal holdout}

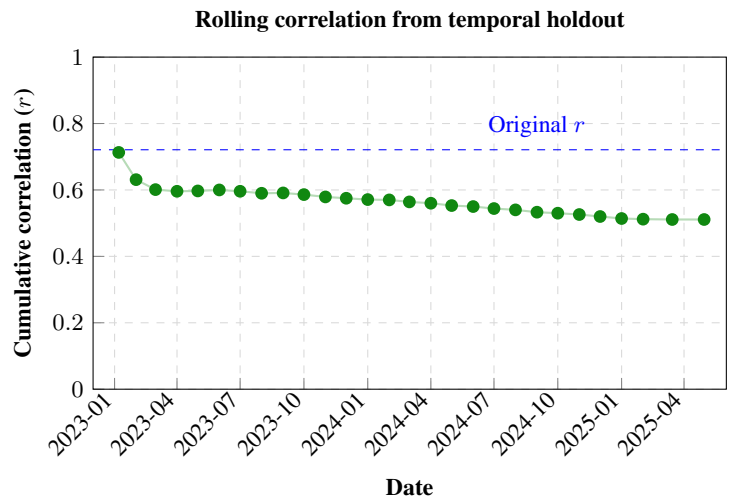
\begin{figure}[!ht]
    \centering
    \begin{tikzpicture}
        \begin{axis}[
            date coordinates in=x,
            xticklabel=\year-\month,
            xticklabel style={
                rotate=45,
                anchor=east
            },
            xtick distance=92,
            xlabel={\textbf{Date}},
            xmin=2022-12-01,
            xmax=2025-06-01,
            xtick={2023-01-01, 2023-04-01, 2023-07-01, 2023-10-01, 2024-01-01, 2024-04-01, 2024-07-01, 2024-10-01, 2025-01-01, 2025-04-01},
            xticklabel={\year-\month},
            ylabel={\textbf{Cumulative correlation ($r$)}},
            ymin=0,
            ymax=1,
            title={\textbf{Rolling correlation from temporal holdout}}
        ]

            \addplot[color=color1!30, mark=*, mark options={color=color1}, thick] coordinates {
                (2023-01-07, 0.713)
                (2023-02-01, 0.631)
                (2023-03-01, 0.601)
                (2023-04-01, 0.596)
                (2023-05-01, 0.597)
                (2023-06-01, 0.600)
                (2023-07-01, 0.596)
                (2023-08-01, 0.590)
                (2023-09-01, 0.591)
                (2023-10-01, 0.586)
                (2023-11-01, 0.579)
                (2023-12-01, 0.575)
                (2024-01-01, 0.571)
                (2024-02-01, 0.570)
                (2024-03-01, 0.564)
                (2024-04-01, 0.560)
                (2024-05-01, 0.553)
                (2024-06-01, 0.550)
                (2024-07-01, 0.544)
                (2024-08-01, 0.540)
                (2024-09-01, 0.533)
                (2024-10-01, 0.530)
                (2024-11-01, 0.526)
                (2024-12-01, 0.520)
                (2025-01-01, 0.514)
                (2025-02-01, 0.512)
                (2025-03-15, 0.511)
                (2025-04-30, 0.511)
            };

            \addplot[dashed, color=color2] coordinates {
                (2022-12-01, 0.721)
                (2025-06-01, 0.721)
            };

            \node at (axis cs:2024-09-01, 0.73) [anchor=south, color=color2] {Original $r$};
            
        \end{axis}
    \end{tikzpicture}

    \caption{This graph visualizes the rolling test correlation of the Bloom-560m model, evaluated with temporal holdout. For example, the point with $x$ value 2024-10 and $y$ value 0.530 represents a 0.530 Pearson correlation between the true and predicted target values between 2023-01 and 2024-10. The blue line denotes the original $r$ found during the scaling-law evaluation of Bloom-560m.}
    \label{fig:temporal-holdout}
\end{figure}

Figure~\ref{fig:temporal-holdout} charts model performance out-of-distribution from January 2023 through April 2025. The initial correlation of $r = 0.713$ closely matched the static test performance of $r = 0.721$ reported during our scaling-law analysis. Performance dropped to $r = 0.631$ after the first month, and then steadily decayed to $r = 0.511$ by mid-2025.

\FloatBarrier

\section{Discussion}
\subsection{Interpretation of experimental findings}

\paragraph{Comparison to prior work.} Our best models represent a substantial advancement over previous state-of-the-art techniques. SChuBERT and MultiSChuBERT report coefficients of determination $R^2 = 0.4$ and $R^2 = 0.454$ respectively, despite the latter using the additional visual information embedded in the figures and layout of the articles \citep{van_Dongen_2020, wenniger2023multischubert}. In comparison, we record a coefficient of determination $R^2 = 0.706$, given only textual information. Current state-of-the-art models such as CiMaTe achieve a Spearman rank correlation of $\rho = 0.436$ in computational linguistics and biological domains \citep{hirako2024cimatecitationcountprediction}, whereas \citet{li-etal-2019-neural} record $\rho = 0.556$ when given the additional use of peer review text. Our Qwen2.5-14B model trained on 16\% of our corpus obtained a Spearman correlation of $\rho = 0.826$, marking a 27-point improvement. 

\paragraph{Scaling-laws and extrapolation.} Our empirical findings underscore both the current limitations and future promise of pre-trained language models for citation rate prediction. First, scaling-law analysis reveals a clear, monotonic improvement across all metrics as data volume increases and a similar (albeit more aberrant) increase as model size increases. We hypothesize that the deviation in performance of mid-sized models is due to their specialized fine-tuning, as both Phi-4-mini-instruct and Qwen2.5-Math-7B were fine-tuned prior to our training regime. We believe this accounts for the deviations reported due to the nature of the fine-tuning performed: Phi-4 was fine-tuned for conversation, whereas the Qwen2.5-7B model was fine-tuned for mathematical reasoning, which we would expect to exhibit more forgetful tendencies of prior knowledge than chat fine-tuning. This is support by our findings in Appendix~\ref{app:reasoning}, where we compared the scaling-laws of DeepSeek's reasoning distilled (fine-tuned) models to that of their original Qwen2.5 counterparts and found a consistent, albeit minor, drop in performance across data volumes and model sizes.

\paragraph{Gradient saliency analysis.} Gradient saliency analysis reveals dominant attribution in title and abstract tokens, with minimal weight attributed to the main content of manuscripts. This is, perhaps, to be expected; titles and abstracts are among the only textual features which are consistently readily available to researchers, not hidden behind paywalls or required downloads. Notably, training without titles/abstracts yields only minor drops in performance ($\Delta r < 0.02$, see Appendix \ref{app:ablations}), indicating these sections serve as convenient surrogates rather than strictly necessary features. This is optimistic for future work, perhaps suggesting that these models are in fact capable of using the content of the research itself to predict citations, rather than simply reward hacking.

\paragraph{Temporal holdout.} Temporal holdout experiments demonstrate the robustness of our approach. Our smallest model maintains performance for up to a month out-of-distribution, and continues to perform better than many previous works up to two years later. The Bloom-560m model was released in 2022, prior to our cutoff date, which verifies that our model's success has not been due to data leakage during pre-training; these models are not memorizing content, they are learning to extract and quantify highly-citable features in academic texts. The degradation in performance after one month out-of-distribution indicates the potential benefit of an online-learning regime in practical deployments of these models. For more details on the potential effects of pre-training see Appendix~\ref{app:pre-training}.

\subsection{Ethical considerations}

The potential misuse of citation-prediction systems is not a subject to be flippantly dismissed. Such tools could facilitate superficial "citation hacking", whereby authors tailor linguistic style to game performance metrics, rather than improving the inherent scientific value of their work. Moreover, it is reasonable to expect such tools to find use in publications and journals as a means of automatic sorting, efficiently discriminating between highly citable papers and their less citable counterparts. However, historical precedents should serve to caution against such a system. For example, Alan Turing's 1936 insights on computability introduced theory that would later become the foundation of modern computer science \citep{turing1936computable}. His work remained almost entirely un-cited for decades and would surely qualify for rejection by an automated citation-prediction based system such as the one we described above, despite its paradigm-shifting effect. It is the belief of the authors that citation-prediction tools should complement, rather than supplant, human judgment, leaving room for the paradigm-shifting works of our time.

\subsection{Limitations \& future work}

Due to limited time and compute, our study was constrained by single-run experiments conducted under a strict seven-day execution window, which inherently restricted the scope and depth of our analysis. This computational budget also necessitated the manual hyperparameter tuning of our framework, which almost certainly resulted in a suboptimal configuration. During this process our models exhibited signs that the initial training phase of the linear head constrained the subsequent full-model fine-tuning to an inferior local minima, a symptom which could likely be mitigated by a more gradual unfreezing of model weights \citep{howard2018universallanguagemodelfinetuning}. Moreover, our gradient saliency analysis and ablations both suggest an unnecessary reliance on peripheral elements during inference. Finally, our models were inherently limited by their training data, which comprised exclusively of biomedical research written in English.

To address these shortcomings, future work may include:
\begin{enumerate}
    \item The integration of \textbf{multi-modal representations} to more comprehensively model scientific manuscripts.
    \item The implementation of an aggressive \textbf{section-wise dropout} during training to compel models to recognize weak-but-valuable signals, thereby increasing robustness.
    \item Leveraging \textbf{progressive training regimes} to prevent trapping the models in suboptimal minima.
    \item Expanding evaluations to include \textbf{non-biomedical and non-English corpora} to verify the generalizability and real-world applicability of direct LLM-based citation prediction.
\end{enumerate}

\section{Conclusion}

We have introduced ForeCite, a unified framework for adapting pre-trained causal language models for direct, end-to-end prediction of average monthly citation rates using only in-text information. Through comprehensive scaling-law experiments, we demonstrated that model performance grows predictably with both parameter count and data volume, and our largest configuration\textemdash Qwen2.5-14B on 16\% of our 900K article corpus\textemdash achieves a Spearman rank correlation of $\rho = 0.826$ on held-out test sets, a 27-point improvement over the previous state-of-the-art. Extrapolations using a bounded least squares model suggest that state-of-the-art LLMs trained on a full dataset could push the performance of our framework significantly farther.

Complementary studies revealed that specialized reasoning fine-tuning does not enhance citation forecasting and may in fact degrade the semantic signals used for this task. Gradient saliency analysis and targeted ablations displayed heavy reliance on titles and abstracts\textemdash convenient shortcuts rather than indispensable features\textemdash highlighting the need for more intelligent training strategies which encourage full text comprehension. 

Altogether, these findings chart a clear path forward: richer multi-modal integration, aggressive dropout schemes to encourage full-text comprehension, and robust evaluation corpora to demonstrate generalizability. While substantial gains remain to be realized at larger scales, ForeCite lays the groundwork for AI-driven research tools capable of guiding academics toward higher-impact writing and, in the long-term, accelerated scientific discovery.

\begin{ack}
    This research was supported by an NSERC Undergraduate Student Research Award (USRA), the NSERC Discovery Grant program and the Canada Research Chairs program. Our deepest thanks to Dr. Shane Arsenault for kindly allowing his work to be featured in our gradient saliency analysis. Our experiments were made possible by the exceptional support team and world-class infrastructure of Compute Canada. We extend a special thanks to Elsevier for granting us access to their publication's data; without their cooperation, this work could not have been realized. Finally, we thank Peter and Dhiraj for their valuable contributions to early discussions.
\end{ack}

\section*{Author contributions}

The roles of each author are described using the CRediT (Contributor Roles Taxonomy) as follows:\\
\textbf{\anon{Gavin Hull}{Author A}}: Conceptualization; Methodology; Software; Validation; Formal analysis; Investigation; Data curation; Writing - original draft; Writing - review \& editing; Visualization; Project administration; Funding acquisition.\\
\textbf{\anon{Dr.\ Alex Bihlo}{Author B}}: Conceptualization; Resources; Writing - review \& editing; Supervision; Project administration; Funding acquisition.

\bibliographystyle{abbrvnat}
\bibliography{refs}

\newpage

\appendix

\section{Hyperparameters}\label{app:hyperparams}

\begin{figure}[!ht]
    \centering
    \textbf{Hyperparameters}\par\medskip

    \begin{tabular}{@{} l l >{\raggedright\arraybackslash}p{5em} @{}}
        \toprule
        \multicolumn{2}{c}{\textbf{Group / Parameter}} & \textbf{Value}\\
        \cmidrule(r){1-2} \cmidrule(l){3-3}

        \textbf{Training}    & Optimizer               & AdamW  \\
                             & Learning Rate           & 1.0e-4 \\
                             & Learning Rate Scheduler & Cosine \\
                             & Weight Decay            & 1.0e-2 \\
                             & Grad. Accum. Steps      & 4      \\
                             & Batch Size              & 2      \\
                             & Epochs                  & 1      \\
        \midrule
        
        \textbf{Fine-tuning} & Optimizer               & AdamW  \\
                             & Learning Rate           & 1.0e-4 \\
                             & Learning Rate Scheduler & Cosine \\
                             & Weight Decay            & 1.0e-2 \\
                             & Grad. Accum. Steps      & 16     \\
                             & Batch Size              & 1      \\
                             & Epochs                  & 3      \\
        \midrule
        
        \textbf{QLoRA}       & Dropout                 & 5.0e-2 \\
                             & Alpha                   & 8      \\
                             & Rank                    & 4      \\
        \bottomrule
    \end{tabular}

    \caption{A table of hyperparameters used for all experiments. Any omitted parameters were left at their default values.}
    \label{tab:hyperparams}

\end{figure}

\FloatBarrier

\section{Model variants}\label{app:models}

\begin{figure}[!ht]
    \centering
    \begin{tabular}{lcc}
        \toprule
        \textbf{Model} & \textbf{Publisher} & \textbf{Parameters (B)} \\
        \midrule

        Bloom-560m                   & BigScience         & 0.56 \\
        Llama-3.2-1B                 & Meta               & 1.24  \\
        Phi-4-mini-instruct          & Microsoft          & 3.84  \\
        Qwen2.5-Math-7B              & Alibaba            & 7.62  \\
        Qwen2.5-14B                  & Alibaba            & 14.8  \\
        DeepSeek-R1-Distill-Qwen-7B  & DeepSeek           & 7.62  \\
        DeepSeek-R1-Distill-Qwen-14B & DeepSeek           & 14.8  \\

        \bottomrule
    \end{tabular}

    \caption{This table contains every base language model used throughout our research, as well as their parameter counts and publishers.}
    \label{tab:models}
\end{figure}

\FloatBarrier

\newpage
\section{Evaluation metrics}\label{app:metrics}

To evaluate model performance, we reported the following metrics:
\begin{itemize}
    \item \textbf{Pearson correlation coefficient ($r$):} Measures the linear relationship between predicted and true (transformed) citation rates. This is our most commonly reported statistic due to its ubiquity and simplicity.
    \item \textbf{Spearman rank correlation ($\rho$):} Captures the monotonic relationship between predicted and true (transformed) citation rates. Equivalent to the Pearson correlation between the rank values of the two variables. This is the statistic we most commonly compare to prior works, as it is invariant under any monotonic transformation, ensuring compatibility between reportings.
    \item \textbf{Coefficient of determination ($R^2$):} Quantifies the proportion of the variation in the true (transformed) citation rates which is explainable by the predicted citation rates, and thus our models.
    \item \textbf{Mean absolute error (MAE):} Represents the average magnitude of the prediction error.
    \item \textbf{Mean squared error (MSE):} Similar to MAE, except it penalizes large deviations more heavily.
\end{itemize}

\clearpage

\begin{figure}[!ht]
    \section{Scaling-laws}\label{app:scaling}
    \centering
    \textbf{Scaling-laws}\par\medskip

    \small
    \begin{tabular}{@{} l c l c c c c c c @{}}
        \toprule
        \textbf{Model name} & \textbf{Parameters} & \textbf{Split} & \textbf{Data \%} & $r \ (\uparrow)$ & $\rho \ (\uparrow)$ & $R^2 \ (\uparrow)$ & \textbf{MSE} $(\downarrow)$ & \textbf{MAE} $(\downarrow)$ \\
        \midrule

        \multirow{10}{*}{Bloom-560m} & \multirow{10}{*}{0.559} & \multirow{5}{*}{train} & 1 & 0.502 & 0.473 & 0.251 & 0.749 & 0.677\\
         & & & 2 & 0.565 & 0.539 & 0.318 & 0.682 & 0.645\\
         & & & 4 & 0.660 & 0.633 & 0.433 & 0.567 & 0.588\\
         & & & 8 & 0.704 & 0.673 & 0.494 & 0.506 & 0.556\\
         & & & 16 & 0.749 & 0.716 & 0.560 & 0.440 & 0.519\\
         \cmidrule(l){3-9}
         & & \multirow{5}{*}{test} & 1 & 0.430 & 0.380 & 0.181 & 1.342 & 0.911\\
         & & & 2 & 0.561 & 0.546 & 0.313 & 1.064 & 0.795\\
         & & & 4 & 0.609 & 0.579 & 0.367 & 0.992 & 0.770\\
         & & & 8 & 0.680 & 0.652 & 0.459 & 0.841 & 0.714\\
         & & & 16 & 0.721 & 0.685 & 0.518 & 0.745 & 0.674\\
         \midrule
        
        \multirow{10}{*}{Llama-3.2-1B} & \multirow{10}{*}{1.24} & \multirow{5}{*}{train} & 1 & 0.719 & 0.688 & 0.516 & 0.484 & 0.544\\
         & & & 2 & 0.794 & 0.765 & 0.630 & 0.370 & 0.476\\
         & & & 4 & 0.824 & 0.800 & 0.677 & 0.323 & 0.443\\
         & & & 8 & 0.848 & 0.824 & 0.718 & 0.282 & 0.414\\
         & & & 16 & 0.862 & 0.839 & 0.742 & 0.258 & 0.396\\
         \cmidrule(l){3-9}
         & & \multirow{5}{*}{test} & 1 & 0.674 & 0.619 & 0.453 & 0.902 & 0.740\\
         & & & 2 & 0.720 & 0.682 & 0.511 & 0.702 & 0.652\\
         & & & 4 & 0.741 & 0.717 & 0.544 & 0.665 & 0.637\\
         & & & 8 & 0.777 & 0.742 & 0.599 & 0.609 & 0.612\\
         & & & 16 & 0.787 & 0.758 & 0.615 & 0.596 & 0.603\\
         \midrule
         
        \multirow{10}{*}{Phi-4-mini-instruct} & \multirow{10}{*}{3.84} & \multirow{5}{*}{train} & 1 & 0.763 & 0.738 & 0.582 & 0.418 & 0.506\\
         & & & 2 & 0.804 & 0.774 & 0.646 & 0.354 & 0.465\\
         & & & 4 & 0.843 & 0.817 & 0.711 & 0.289 & 0.421\\
         & & & 8 & 0.865 & 0.844 & 0.747 & 0.253 & 0.391\\
         & & & 16 & 0.872 & 0.852 & 0.760 & 0.240 & 0.381\\
         \cmidrule(l){3-9}
         & & \multirow{5}{*}{test} & 1 & 0.657 & 0.630 & 0.423 & 0.876 & 0.734\\
         & & & 2 & 0.723 & 0.698 & 0.519 & 0.727 & 0.659\\
         & & & 4 & 0.752 & 0.727 & 0.559 & 0.623 & 0.613\\
         & & & 8 & 0.789 & 0.767 & 0.619 & 0.578 & 0.592\\
         & & & 16 & 0.802 & 0.781 & 0.641 & 0.549 & 0.575\\
         \midrule
        
        \multirow{10}{*}{Qwen2.5-Math-7B} & \multirow{10}{*}{7.62} & \multirow{5}{*}{train} & 1 & 0.759 & 0.728 & 0.575 & 0.425 & 0.507\\
         & & & 2 & 0.810 & 0.780 & 0.656 & 0.344 & 0.459\\
         & & & 4 & 0.852 & 0.826 & 0.725 & 0.275 & 0.410\\
         & & & 8 & 0.873 & 0.851 & 0.761 & 0.239 & 0.380\\
         & & & 16 & 0.890 & 0.870 & 0.790 & 0.210 & 0.356\\
         \cmidrule(l){3-9}
         & & \multirow{5}{*}{test} & 1 & 0.645 & 0.607 & 0.411 & 0.880 & 0.736\\
         & & & 2 & 0.721 & 0.696 & 0.517 & 0.752 & 0.678\\
         & & & 4 & 0.753 & 0.724 & 0.563 & 0.648 & 0.626\\
         & & & 8 & 0.776 & 0.750 & 0.598 & 0.625 & 0.615\\
         & & & 16 & 0.786 & 0.758 & 0.613 & 0.600 & 0.602\\
         \midrule
        
        \multirow{10}{*}{Qwen2.5-14B} & \multirow{10}{*}{14.8} & \multirow{5}{*}{train} & 1 & 0.811 & 0.791 & 0.654 & 0.346 & 0.454\\
         & & & 2 & 0.866 & 0.848 & 0.746 & 0.254 & 0.388\\
         & & & 4 & 0.912 & 0.898 & 0.831 & 0.169 & 0.316\\
         & & & 8 & 0.924 & 0.911 & 0.852 & 0.148 & 0.295\\
         & & & 16 & \textbf{0.936} & \textbf{0.924} & \textbf{0.875} & \textbf{0.125} & \textbf{0.270}\\
         \cmidrule(l){3-9}
         & & \multirow{5}{*}{test} & 1 & 0.752 & 0.754 & 0.561 & 0.653 & 0.615\\
         & & & 2 & 0.797 & 0.779 & 0.632 & 0.544 & 0.575\\
         & & & 4 & 0.826 & 0.797 & 0.679 & 0.487 & 0.544\\
         & & & 8 & 0.832 & 0.809 & 0.687 & 0.472 & 0.534\\
         & & & 16 & \textbf{0.844} & \textbf{0.826 }& \textbf{0.706} & \textbf{0.452} & \textbf{0.522}\\
         \midrule
         
    \end{tabular}

    \caption{Results of scaling-law analysis. Model sizes span 0.5B to 14B parameters and data volumes span between 1\% ($\approx$90K) and 16\% ($\approx$144K). Best performance is found in \textbf{bold} for both train and test splits.}
    \label{tab:scaling}
    
\end{figure}

\FloatBarrier

\section{Scaling-law extrapolation}\label{app:curve-fit}

\begin{figure}[!ht]
    \centering
    \textbf{Least squares fit parameters}\par\medskip

    \begin{tabular}{lccc}
        \toprule
         & $\beta_0$ & $\beta_1$ & $\beta_2$\\
        \midrule
        $r(p, d)$    & 0.6771 & 0.0689 & 0.0767\\
        $\rho(p, d)$ & 0.6260 & 0.0698 & 0.0724\\
        \bottomrule
    \end{tabular}

    \caption{Model parameters derived for scaling-laws using least squares.}
    \label{tab:curve-fit}
\end{figure}

\section{Gradient saliency heatmap}\label{app:gradient-saliency}

\begin{figure}[!ht]
    \centering
    \textbf{Token-wise gradient saliency heatmap}\par\medskip

    \fcolorbox{black}{white}{
    \parbox{\linewidth}{\scriptsize
        \noindent
        \wordheat{-0.08125}{~Multiple}\wordheat{0.29625}{~sclerosis}\wordheat{-0.120625}{~(}\wordheat{-0.0471875}{MS}\wordheat{0.3475}{)}\wordheat{-0.054375}{~is}\wordheat{-0.085625}{~an}\wordheat{0.0675}{~immune}\wordheat{0.038125}{-mediated}\wordheat{-0.01140625}{~dem}\wordheat{-0.00373046875}{y}\wordheat{-0.0428125}{el}\wordheat{0.061875}{inating}\wordheat{0.0825}{~disease}\wordheat{0.0571875}{~of}\wordheat{0.03953125}{~the}\wordheat{0.005390625}{~central}\wordheat{-0.0665625}{~nervous}\wordheat{-0.0040234375}{~system}\wordheat{0.02265625}{~accompanied}\wordheat{0.028125}{~by}\wordheat{-0.00921875}{~chronic}\wordheat{-0.1875}{~inflammation}\wordheat{-0.0709375}{,}\wordheat{0.008359375}{~ax}\wordheat{0.02390625}{onal}\wordheat{0.0440625}{~loss}\wordheat{-0.0721875}{,}\wordheat{-0.07875}{~and}\wordheat{0.03359375}{~neuro}\wordheat{-0.0134375}{de}\wordheat{-0.11125}{generation}\wordheat{-0.27875}{.}\wordheat{0.0346875}{~Trad}\wordheat{-0.2275}{itionally}\wordheat{0.0378125}{,}\wordheat{-0.14125}{~MS}\wordheat{0.06}{~has}\wordheat{0.01828125}{~been}\wordheat{0.058125}{~thought}\wordheat{-0.0092578125}{~of}\wordheat{0.080625}{~as}\wordheat{0.010859375}{~a}\wordheat{0.01640625}{~T}\wordheat{-0.088125}{-cell}\wordheat{-0.01734375}{~mediated}\wordheat{-0.005234375}{~disease}\wordheat{0.0290625}{,}\wordheat{0.17625}{~but}\wordheat{-0.03234375}{~research}\wordheat{-0.128125}{~over}\wordheat{0.001572265625}{~the}\wordheat{0.08375}{~past}\wordheat{0.0734375}{~decade}\wordheat{-0.03953125}{~has}\wordheat{-0.02765625}{~demonstrated}\wordheat{0.010703125}{~the}\wordheat{-0.109375}{~importance}\wordheat{0.001689453125}{~of}\wordheat{-0.06125}{~B}\wordheat{0.0453125}{~cells}\wordheat{0.015703125}{~in}\wordheat{0.002314453125}{~both}\wordheat{-0.02390625}{~acute}\wordheat{-0.0190625}{~dem}\wordheat{-0.0434375}{y}\wordheat{-0.02734375}{el}\wordheat{0.02359375}{ination}\wordheat{0.013828125}{~and}\wordheat{-0.0054296875}{~disease}\wordheat{-0.1675}{~progression}\wordheat{0.014921875}{.}\wordheat{0.13375}{~The}\wordheat{-0.04375}{~highly}\wordheat{0.02015625}{~selective}\wordheat{0.0975}{~irreversible}\wordheat{-0.086875}{~Brut}\wordheat{-0.3025}{on}\wordheat{-0.012734375}{~Ty}\wordheat{-0.0215625}{ros}\wordheat{-0.073125}{ine}\wordheat{-0.0503125}{~Kin}\wordheat{0.005078125}{ase}\wordheat{-0.02203125}{~(}\wordheat{-0.0076953125}{B}\wordheat{-1.0}{TK}\wordheat{0.0484375}{)}\wordheat{0.14125}{~inhibitors}\wordheat{-0.0615625}{~ev}\wordheat{0.18}{ob}\wordheat{-0.93}{rut}\wordheat{-0.159375}{in}\wordheat{-0.68}{ib}\wordheat{-0.0615625}{,}\wordheat{0.0171875}{~to}\wordheat{0.0353125}{le}\wordheat{0.09625}{br}\wordheat{-0.3275}{utin}\wordheat{-0.21125}{ib}\wordheat{-0.17625}{,}\wordheat{0.01609375}{~and}\wordheat{-0.0875}{~o}\wordheat{-0.086875}{rel}\wordheat{-0.1075}{ab}\wordheat{0.02}{rut}\wordheat{-0.0575}{in}\wordheat{-0.3475}{ib}\wordheat{-0.0925}{,}\wordheat{-0.15875}{~and}\wordheat{0.0603125}{~the}\wordheat{0.056875}{~reversible}\wordheat{0.0078125}{~B}\wordheat{-0.1925}{TK}\wordheat{0.128125}{~inhibitor}\wordheat{-0.0471875}{~f}\wordheat{0.038125}{ene}\wordheat{0.0490625}{br}\wordheat{-0.0525}{utin}\wordheat{-0.175}{ib}\wordheat{0.049375}{,}\wordheat{0.13625}{~all}\wordheat{-0.15375}{~target}\wordheat{-0.02296875}{~B}\wordheat{0.033125}{-cell}\wordheat{0.080625}{~activation}\wordheat{0.001474609375}{~and}\wordheat{-0.17625}{~aspects}\wordheat{0.089375}{~of}\wordheat{-0.13}{~innate}\wordheat{-0.12125}{~immunity}\wordheat{-0.1375}{,}\wordheat{-0.158125}{~including}\wordheat{-0.001982421875}{~macro}\wordheat{0.0696875}{ph}\wordheat{2.956390380859375e-06}{age}\wordheat{-0.095625}{~and}\wordheat{0.0721875}{~micro}\wordheat{-0.03609375}{gl}\wordheat{-0.0086328125}{ia}\wordheat{-0.076875}{~biology}\wordheat{-0.525}{.}\wordheat{0.0315625}{~The}\wordheat{-0.0051171875}{~c}\wordheat{-0.0103125}{-K}\wordheat{0.156875}{IT}\wordheat{0.159375}{~inhibitor}\wordheat{0.20375}{~mas}\wordheat{1.0}{itin}\wordheat{-0.131875}{ib}\wordheat{-0.08875}{~mitig}\wordheat{-0.115625}{ates}\wordheat{0.001357421875}{~neuro}\wordheat{0.0072265625}{in}\wordheat{-0.049375}{flamm}\wordheat{0.019140625}{ation}\wordheat{-0.31875}{~by}\wordheat{-0.0278125}{~controlling}\wordheat{-0.011328125}{~the}\wordheat{0.01984375}{~survival}\wordheat{-0.0546875}{,}\wordheat{-0.0165625}{~migration}\wordheat{0.0590625}{,}\wordheat{-0.11875}{~and}\wordheat{0.018046875}{~de}\wordheat{-0.1925}{gran}\wordheat{0.00396484375}{ulation}\wordheat{0.012890625}{~of}\wordheat{-0.15125}{~mast}\wordheat{-0.27625}{~cells}\wordheat{-0.09}{,}\wordheat{-0.045}{~leading}\wordheat{0.083125}{~to}\wordheat{-0.0090625}{~the}\wordheat{-0.09375}{~inhibition}\wordheat{-0.14375}{~of}\wordheat{-0.020625}{~pro}\wordheat{0.02359375}{in}\wordheat{-0.0228125}{flammatory}\wordheat{-0.02375}{~and}\wordheat{0.0042578125}{~v}\wordheat{0.06375}{aso}\wordheat{0.011171875}{active}\wordheat{-0.01390625}{~molecular}\wordheat{0.1}{~casc}\wordheat{-0.0721875}{ades}\wordheat{0.0008154296875}{~that}\wordheat{0.035}{~result}\wordheat{0.17}{~from}\wordheat{0.083125}{~mast}\wordheat{-0.4975}{~cell}\wordheat{0.0259375}{~activation}\wordheat{-0.26625}{.}\wordheat{-0.38}{~This}\wordheat{-0.4575}{~article}\wordheat{-0.5875}{~will}\wordheat{0.016015625}{~review}\wordheat{-0.375}{~and}\wordheat{-0.090625}{~critically}\wordheat{0.03359375}{~app}\wordheat{0.2}{raise}\wordheat{-0.0571875}{~the}\wordheat{-0.18375}{~ongoing}\wordheat{0.01953125}{~clinical}\wordheat{-0.134375}{~trials}\wordheat{-0.10625}{~of}\wordheat{-0.1225}{~two}\wordheat{0.086875}{~classes}\wordheat{0.015078125}{~of}\wordheat{-0.09375}{~receptor}\wordheat{0.025}{~ty}\wordheat{0.002451171875}{ros}\wordheat{-0.06625}{ine}\wordheat{-0.0403125}{~kinase}\wordheat{0.03171875}{~inhibitors}\wordheat{0.03234375}{~that}\wordheat{0.01015625}{~are}\wordheat{-0.078125}{~emerging}\wordheat{-0.0421875}{~as}\wordheat{-0.22625}{~potential}\wordheat{-0.0278125}{~medical}\wordheat{-0.075625}{~treatments}\wordheat{-0.155625}{~for}\wordheat{0.03828125}{~the}\wordheat{0.075}{~varying}\wordheat{-0.041875}{~sub}\wordheat{-0.025}{types}\wordheat{-0.05125}{~of}\wordheat{-0.18375}{~MS}\wordheat{0.120625}{:}\wordheat{0.001328125}{~B}\wordheat{-0.79}{TK}\wordheat{0.0178125}{~inhibitors}\wordheat{-0.13875}{~and}\wordheat{-0.15}{~c}\wordheat{0.0016796875}{-K}\wordheat{0.13375}{IT}\wordheat{0.02703125}{~inhibitors}\wordheat{-0.23}{.}\wordheat{0.025}{~Specifically}\wordheat{0.104375}{,}\wordheat{-0.0653125}{~this}\wordheat{-0.01609375}{~review}\wordheat{-0.0975}{~will}\wordheat{-0.0390625}{~attempt}\wordheat{0.030625}{~to}\wordheat{0.1275}{~answer}\wordheat{-0.0315625}{~whether}\wordheat{-0.053125}{~B}\wordheat{-0.125625}{TK}\wordheat{0.025}{~inhibitors}\wordheat{-0.00236328125}{~have}\wordheat{0.099375}{~measurable}\wordheat{-0.0578125}{~positive}\wordheat{-0.017734375}{~clinical}\wordheat{0.007421875}{~effects}\wordheat{0.005390625}{~on}\wordheat{0.13}{~patients}\wordheat{-0.03796875}{~with}\wordheat{0.01890625}{~RR}\wordheat{0.12125}{MS}\wordheat{0.018203125}{,}\wordheat{0.010859375}{~SP}\wordheat{-0.144375}{MS}\wordheat{-0.0093359375}{~with}\wordheat{0.004140625}{~rel}\wordheat{0.02390625}{apses}\wordheat{-0.21375}{,}\wordheat{0.0365625}{~rel}\wordheat{-0.0534375}{apse}\wordheat{0.03}{-free}\wordheat{-0.011484375}{~SP}\wordheat{-0.0515625}{MS}\wordheat{-0.1925}{,}\wordheat{-0.075625}{~and}\wordheat{0.02578125}{~PP}\wordheat{-0.002373046875}{MS}\wordheat{-0.01203125}{~through}\wordheat{-0.05625}{~their}\wordheat{-0.00142578125}{~effect}\wordheat{0.02328125}{~on}\wordheat{0.064375}{~MRI}\wordheat{0.00045654296875}{~T}\wordheat{-0.075}{1}\wordheat{0.01828125}{~lesions}\wordheat{0.0525}{;}\wordheat{-0.0309375}{~annual}\wordheat{0.0553125}{ized}\wordheat{-0.00310546875}{~rel}\wordheat{0.038125}{apse}\wordheat{0.00486328125}{~rate}\wordheat{-0.002060546875}{;}\wordheat{0.0515625}{~ED}\wordheat{-0.4225}{SS}\wordheat{-0.143125}{~scale}\wordheat{0.04875}{;}\wordheat{-0.08625}{~MS}\wordheat{0.0575}{FC}\wordheat{-0.133125}{~score}\wordheat{-0.090625}{;}\wordheat{0.017265625}{~and}\wordheat{0.01828125}{~time}\wordheat{-0.01171875}{~to}\wordheat{-0.06625}{~onset}\wordheat{0.00640625}{~of}\wordheat{-0.128125}{~composite}\wordheat{0.01203125}{~}\wordheat{0.01140625}{12}\wordheat{-0.03}{-week}\wordheat{0.03765625}{~confirmed}\wordheat{0.055625}{~disability}\wordheat{0.0203125}{~progression}\wordheat{-0.5875}{.}\wordheat{0.22125}{~Additionally}\wordheat{0.17}{,}\wordheat{0.040625}{~this}\wordheat{0.034375}{~review}\wordheat{0.069375}{~will}\wordheat{0.10625}{~examine}\wordheat{0.02453125}{~the}\wordheat{0.24875}{~literature}\wordheat{-0.0615625}{~to}\wordheat{0.041875}{~determine}\wordheat{0.0709375}{~if}\wordheat{0.185}{~mas}\wordheat{0.435}{itin}\wordheat{-0.395}{ib}\wordheat{0.02796875}{~has}\wordheat{0.01921875}{~positive}\wordheat{0.014609375}{~clinical}\wordheat{-0.0734375}{~effects}\wordheat{-0.05625}{~on}\wordheat{-0.0415625}{~patients}\wordheat{-0.0075}{~with}\wordheat{-0.21375}{~PP}\wordheat{-0.845}{MS}\wordheat{-0.0303125}{~or}\wordheat{0.001455078125}{~rel}\wordheat{-0.03609375}{apse}\wordheat{0.0246875}{-free}\wordheat{-0.0203125}{~SP}\wordheat{0.03921875}{MS}\wordheat{-0.0428125}{~through}\wordheat{0.013671875}{~its}\wordheat{0.00287109375}{~effect}\wordheat{0.0059765625}{~on}\wordheat{-0.0709375}{~ED}\wordheat{0.16875}{SS}\wordheat{-0.03390625}{~or}\wordheat{0.02296875}{~MS}\wordheat{-0.011796875}{FC}\wordheat{-0.11625}{~scores}\wordheat{0.22125}{.}
    }
    }

    \begin{tikzpicture}
        \begin{axis}[
            hide axis,
            scale only axis,
            xmin=-1,
            xmax=1,
            ymin=-1,
            ymax=1,
            width=7cm, height=0.25cm,
            colormap={rgwhite}{rgb255(0cm)=(230,25,75) rgb255(1cm)=(255,255,255) rgb255(2cm)=(60,180,75)},
            colorbar horizontal,
            point meta min=-1,
            point meta max=1,
            colorbar style={
                tick align=outside,
                xtick={-1, 0, 1},
                xticklabels={Minimum gradient, 0, Maximum gradient},
                xlabel={\textbf{Scaled gradient value}}
            }
        ]

        \addplot [forget plot] coordinates {(0,0)};
        \end{axis}
    \end{tikzpicture}
    \vspace{1ex}
    
    \caption{A token-level heatmap of model gradients for \citet{arsenault2022does}, computed using Llama-3.2-1B trained on 16\% of our corpus. Red tokens decreased predicted citation rate, whereas green tokens increased the predicted citation rate. Color intensity is proportional to gradient magnitude, reflecting intensity of impact on predicted citation rate.}
    \label{fig:gradient-saliency}
\end{figure}
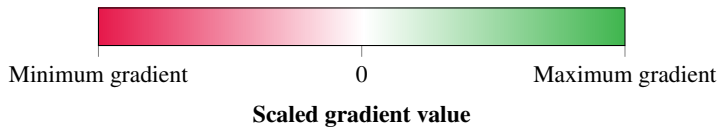

\FloatBarrier

\section{Large reasoning models}\label{app:reasoning}

\begin{figure}[!ht]
    \centering
    \begin{tikzpicture}
        \begin{axis}[
            xlabel={\textbf{Data volume (\%)}},
            xmode=log,
            log basis x=2,
            xtick={0.01,0.02,0.04,0.08,0.16},
            xticklabels={1,2,4,8,16},
            ylabel={\textbf{Test correlation ($r$)}},
            legend pos=south east,
            legend style={
                nodes={
                    scale=0.8,
                    transform shape
                }
            },
            legend cell align=left,
            title={\textbf{LLMs vs. LRMs}}
        ]

            \addplot[color=color1!30, mark=pentagon*, mark options={color=color1}, thick] coordinates {
                (0.01, 0.7552)
                (0.02, 0.7973)
                (0.04, 0.8270)
                (0.08, 0.8335)
                (0.16, 0.8441)
            };
            \addlegendentry{Qwen2.5-14B}

            \addplot[color=color1!30, mark=pentagon*, mark options={color=color1, solid}, thick, dashed] coordinates {
                (0.01, 0.7615)
                (0.02, 0.7988)
                (0.04, 0.8093)
                (0.08, 0.8190)
                (0.16, 0.8297)
            };
            \addlegendentry{DeepSeek-R1-Distill-Qwen-14B}

            \addplot[color=color2!30, mark=triangle*, mark options={color=color2}, thick] coordinates {
                (0.01, 0.6463)
                (0.02, 0.7230)
                (0.04, 0.7540)
                (0.08, 0.7776)
                (0.16, 0.7884)
            };
            \addlegendentry{Qwen2.5-Math-7B}

            \addplot[color=color2!30, mark=triangle*, mark options={color=color2, solid}, thick, dashed] coordinates {
                (0.01, 0.6826)
                (0.02, 0.7025)
                (0.04, 0.7358)
                (0.08, 0.7679)
                (0.16, 0.7762)
            };
            \addlegendentry{DeepSeek-R1-Distill-Qwen-7B}
            
        \end{axis}
    \end{tikzpicture}

    \caption{Effect of reasoning fine-tuning on citation prediction performance as measured in Pearson correlation at varying degrees of data availability.}
    \label{fig:reasoning}
\end{figure}
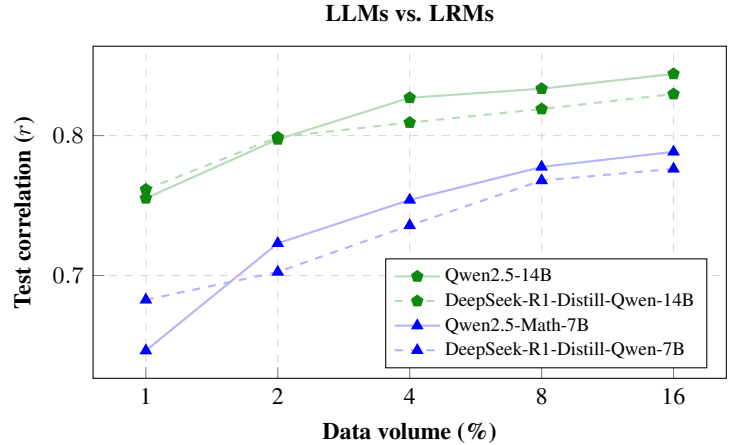

We examined the effect of specialized reasoning fine-tuning on citation prediction by comparing each Qwen2.5 checkpoint to its DeepSeek distilled counterpart. Figure~\ref{fig:reasoning} plots Pearson correlation for the original and distilled Qwen2.5 models at increasing data fractions. At each evaluation of 4\%+ data volume, both Qwen2.5 variants surpassed their DeepSeek counterparts, indicating that reasoning fine-tuning does not improve, and may even slightly reduce, citation prediction performance.

This suggests that the latent semantic knowledge stored inside the LLMs which is most useful during citation prediction is contextual information rather than logical information, as we would expect contextual information accumulated during training to deteriorate during fine-tuning of any kind. This theory is consistent with the deviation in performance reported from the Phi-4-mini-instruct and Qwen2.5-Math-7B models during scaling-law analysis. We suspect this trend may reverse as more "intelligent" models begin to use the main content of articles for their predictions rather than relying on title and abstract tokens.

\FloatBarrier

\section{Ablations}\label{app:ablations}

To assess the aforementioned reliance on peripheral elements in sample articles, we performed two targeted ablation studies on Llama-3.2-1B (16\% data): in one we removed all title tokens and in the other we removed all abstract tokens. 

\begin{figure}[!ht]
    \centering
    \begin{tikzpicture}
        \begin{axis}[
            ybar,
            bar width=30pt,
            ylabel={\textbf{Test correlation ($r$)}},
            ymin=0.7,
            ymax=0.9,
            symbolic x coords={Train,Test},
            xtick=data,
            enlarge x limits=0.5,
            nodes near coords,
            legend style={
                at={(0.5, -0.15)},
                anchor=north,
                legend columns=-1
            },
            legend cell align=left,
            title={\textbf{Impact of titles/abstracts on performance}}
        ]

        \addplot+[
                draw=color1,
                fill=color1!30,
                nodes near coords,
                every node near coord/.append style={text=color1, /pgf/number format/.cd, fixed, precision=3},
            ] coordinates {
                (Train,0.862)
                (Test,0.787)
            };
            
            \addplot+[
                draw=color2,
                fill=color2!30,
                nodes near coords,
                every node near coord/.append style={text=color2, /pgf/number format/.cd, fixed, precision=3},
            ] coordinates {
                (Train,0.851)
                (Test,0.782)
            };

            \addplot+[
                draw=color3,
                fill=color3!30,
                nodes near coords,
                every node near coord/.append style={text=color3, /pgf/number format/.cd, fixed, precision=3},
            ] coordinates {
                (Train,0.853)
                (Test,0.777)
            };

            \legend{Original, No titles, No abstracts}
            
        \end{axis}
    \end{tikzpicture}

    \caption{Pearson correlation for full text vs. no-title vs. no-abstract ablations.}
    \label{fig:ablations}
\end{figure}
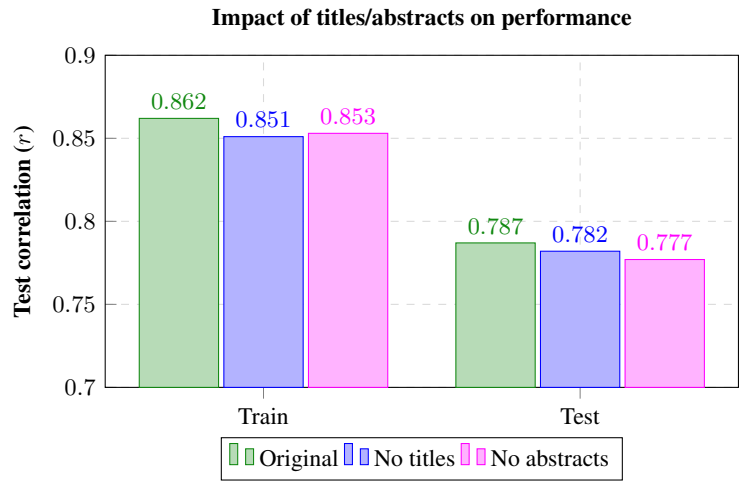

Figure~\ref{fig:ablations} displays the results: the full model achieved $r = 0.787$, the no-title variant $r = 0.782$, and the no-abstract variant $r = 0.777$. These relatively negligible drops suggest that the prior reliance on abstract and title tokens may have been a reflection of convenience rather than necessity. This may motivate the use of section-by-section dropout schemes in future work to encourage more robust predictions.

\FloatBarrier

\section{Effect of pre-training}\label{app:pre-training}

In order to verify the utility of pre-training in our predictions, we evaluated Llama-3.2-1B on various data volumes after pre-training (the original model), and after a random re-initialization (simulating the model prior to pre-training). Figure~\ref{app:pre-training} demonstrates a steep drop in performance after re-initialization across all data volumes. The observed performance drop narrowed from $\Delta r = 0.617$ to $\Delta r = 0.313$ as data volume increased.

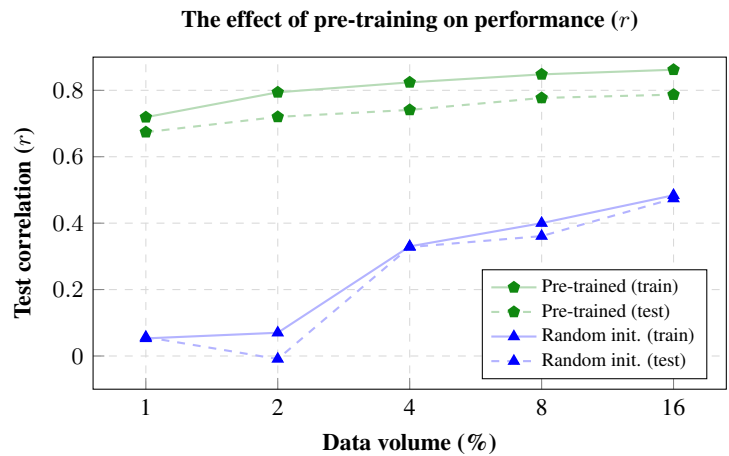
\begin{figure}[!ht]
    \centering
    \begin{tikzpicture}
        \begin{axis}[
            xlabel={\textbf{Data volume (\%)}},
            xmode=log,
            log basis x=2,
            xtick={0.01,0.02,0.04,0.08,0.16},
            xticklabels={1,2,4,8,16},
            ylabel={\textbf{Test correlation ($r$)}},
            ymin=-0.1,
            ymax=0.9,
            legend pos=south east,
            legend style={
                nodes={
                    scale=0.8,
                    transform shape
                }
            },
            legend cell align=left,
            title={\textbf{The effect of pre-training on performance ($r$)}}
        ]

            \addplot[color=color1!30, mark=pentagon*, mark options={color=color1}, thick] coordinates {
                (0.01, 0.719)
                (0.02, 0.794)
                (0.04, 0.824)
                (0.08, 0.848)
                (0.16, 0.862)
            };
            \addlegendentry{Pre-trained (train)}

            \addplot[color=color1!30, mark=pentagon*, mark options={color=color1, solid}, thick, dashed] coordinates {
                (0.01, 0.674)
                (0.02, 0.720)
                (0.04, 0.741)
                (0.08, 0.777)
                (0.16, 0.787)
            };
            \addlegendentry{Pre-trained (test)}

            \addplot[color=color2!30, mark=triangle*, mark options={color=color2}, thick] coordinates {
                (0.01, 0.053)
                (0.02, 0.070)
                (0.04, 0.330)
                (0.08, 0.400)
                (0.16, 0.484)
            };
            \addlegendentry{Random init. (train)}

            \addplot[color=color2!30, mark=triangle*, mark options={color=color2, solid}, thick, dashed] coordinates {
                (0.01,  0.057)
                (0.02, -0.009)
                (0.04,  0.328)
                (0.08,  0.361)
                (0.16,  0.474)
            };
            \addlegendentry{Random init. (test)}
            
        \end{axis}
    \end{tikzpicture}

    \caption{Effect of pre-training on citation prediction performance as measured in Pearson correlation at varying degrees of data availability.}
    \label{fig:pre-training}
\end{figure}

These results confirm our suspicions that the semantic knowledge learned during pre-training is useful in discerning the features which characterize highly-citable scientific text.

\end{document}